\documentclass[letterpaper]{article} 
\usepackage{aaai25}  
\usepackage{times}  
\usepackage{helvet}  
\usepackage{courier}  
\usepackage[hyphens]{url}  
\usepackage{graphicx} 

\usepackage{multirow}
\usepackage{booktabs}
\usepackage{amsmath,amssymb,amsfonts}
\usepackage{color,xcolor}
\usepackage{subfigure}

\usepackage{graphicx} 
\usepackage{natbib}  
\usepackage{caption} 
\usepackage{algorithm}
\usepackage{algorithmic}

\usepackage{newfloat}
\usepackage{listings}
\DeclareCaptionStyle{ruled}{labelfont=normalfont,labelsep=colon,strut=off} 
\floatstyle{ruled}
\newfloat{listing}{tb}{lst}{}
\floatname{listing}{Listing}
\title{CognTKE: A Cognitive Temporal Knowledge Extrapolation  Framework}
\author{
    Wei Chen\textsuperscript{\rm 1,3}, Yuting Wu\textsuperscript{\rm 2}, Shuhan Wu\textsuperscript{\rm 1,3}, Zhiyu Zhang\textsuperscript{\rm 1,3}, Mengqi Liao\textsuperscript{\rm 1,3}, \\
    Youfang Lin\textsuperscript{\rm 1,3}, Huaiyu Wan\textsuperscript{\rm 1,3} \thanks{Corresponding authors.}
}
\affiliations{
    \textsuperscript{\rm 1}School of Computer Science \& Technology, Beijing Jiaotong University, China\\
    \textsuperscript{\rm 2}School of Software, Beijing Jiaotong University, China\\
    \textsuperscript{\rm 3}Beijing Key Laboratory of Traffic Data Mining and Embodied Intelligence, Beijing, China\\

       \{w\_chen, ytwu1, wushuhan, zyuzhang, mqliao, yflin, hywan\}@bjtu.edu.cn

}

\usepackage{bibentry}

\begin{document}

\maketitle

\begin{abstract}
Reasoning future unknowable facts on temporal knowledge graphs (TKGs) is a challenging task, holding significant academic and practical values for various fields. 
Existing studies exploring explainable reasoning concentrate on modeling comprehensible temporal paths relevant to the query. Yet, these path-based methods primarily focus on local temporal paths appearing in recent times, failing to capture the complex temporal paths in TKG and resulting in the loss of longer historical relations related to the query. 
Motivated by the Dual Process Theory in cognitive science, we propose a \textbf{Cogn}itive \textbf{T}emporal \textbf{K}nowledge \textbf{E}xtrapolation framework (CognTKE), which introduces a novel temporal cognitive relation directed graph (TCR-Digraph) and performs interpretable global shallow reasoning and local deep reasoning over the TCR-Digraph. Specifically, the proposed TCR-Digraph is constituted by retrieving significant local and global historical temporal relation paths associated with the query. In addition, CognTKE presents the global shallow reasoner and the local deep reasoner to perform global one-hop temporal relation reasoning (System 1) and local complex multi-hop path reasoning (System 2) over the TCR-Digraph, respectively. 
The experimental results on four benchmark datasets demonstrate that CognTKE achieves significant improvement in accuracy compared to the state-of-the-art baselines and delivers excellent zero-shot reasoning ability. \textit{The code is available at https://github.com/WeiChen3690/CognTKE}.
\end{abstract}

%

\section{Introduction}
\begin{figure}[htbp]
	\setlength{\abovecaptionskip}{1pt}
	\centering
	\includegraphics[width=0.95\linewidth]{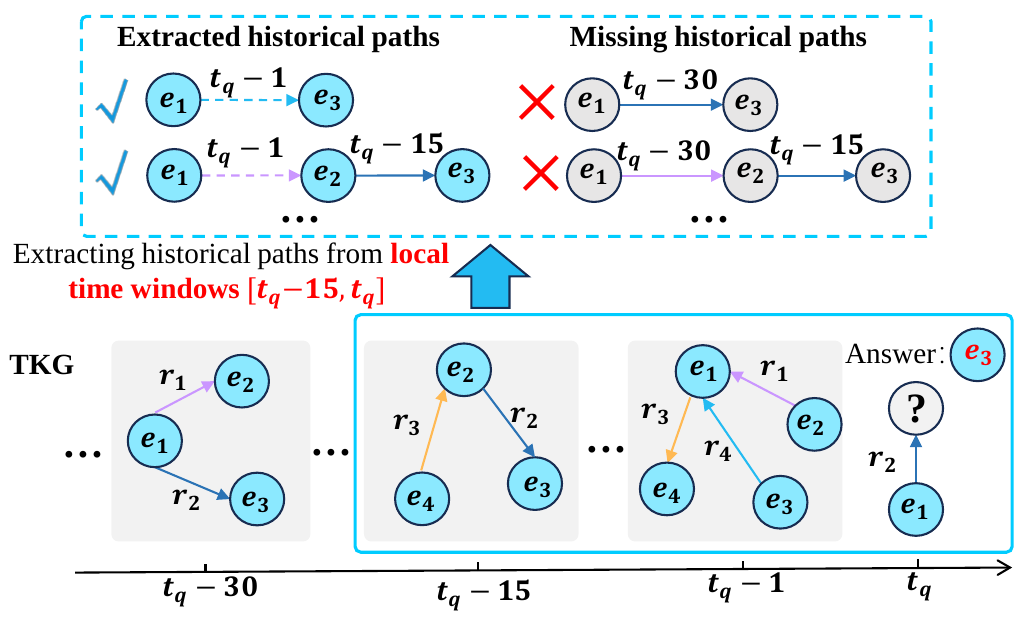}
	\caption{An example illustrates longer history temporal relation path loss. Relations are denoted using solid lines with different colors, and inverse relations are denoted using dashed lines. Black solid line denotes the identity relation.}
	\label{fig:Challeng}
\end{figure}
Temporal knowledge graphs (TKGs) store the structured dynamic facts in the real world, consisting of a sequence of knowledge graph (KG) snapshots accompanied by corresponding timestamps. Each fact in TKGs can be represented as a quadruple in the form of (subject, relation, object, time), e.g., \textit{(The World Cup, Be\_held\_at, Qatar, 2022)}.
Reasoning over TKGs aims to infer new facts based on the historical KG snapshots, which can be divided into two settings: interpolation and extrapolation. According to the given history facts in time interval $[t_0, t_{|\mathcal{T}|}]$,
the former attempts to complete missing facts that appear in the past time $t \in [t_0, t_{|\mathcal{T}|}]$, while the latter focuses on predicting facts (events) happening in the future time $t > t_{|\mathcal{T}|}$.  
Due to its ability to infer future new events, extrapolation in TKG has a wider range of application scenarios \cite{ChenWGHZLLL22,Event}, such as stock prediction, traffic prediction, and medical assistance. This paper primarily emphasizes the more challenging  TKG extrapolation task that forecasts future unknown facts (or events).	

Many arts for TKG extrapolation have been extensively explored and achieved excellent prediction performance. These studies mainly involve two categories \cite{ALRE_IR, liang2024survey}: embedding-based methods and path-based methods. Embedding-based methods learn the evolution embeddings of entity and relation by modeling the facts repeating and cyclic patterns, such as CyGNet \cite{CyGNet} and CENET \cite{CENET}, and the local adjacent facts evolution patterns, such as RE-GCN \cite{RE-GCN} and RETIA \cite{RETIA}. 
However, due to the black-box characteristics of embedding-based methods, they lack explicit evidence to explain the reasoning results.

Naturally, some path-based methods such as TLogic \cite{Tlogic} and KartGPS \cite{XinC24}, are proposed to better capture the chain of evidence for TKG reasoning. 
Nevertheless,  
path-based methods focus on temporal relational paths modeling in the specific time window closest to the query time, resulting in the loss of longer historical information. As shown in Figure \ref{fig:Challeng}, 
given a local time window $[t_q-15,t_q-1]$, existing path-based methods only capture paths $(e_1, {r_4}^{-1}, e_3, t_{q}-1)$\footnote{${r_4}^{-1}$ denotes the inverse relation of the relation $r_4$} and  $(e_1, {r_1}^{-1}, e_2, t_{q}-1) \rightarrow (e_2, r_2, e_3, t_{q}-15) $ based on the query $(e_1, r_2, ?, t_{q})$, but ignoring distant historical paths information, such as $(e_1, r_1, e_2, t_{q}-30) \rightarrow (e_2, {r_2}, e_3, t_{q}-15)$. Although the time window of path modeling can extend to the entire history, it makes path retrieval expensive and may bring numerous earlier temporal relations irrelevant to the query.
In addition,
the expressiveness of path-based methods is still limited to individual paths, which leads to only a small subset of first-order logical formulas being represented in TKGs.
Subgraphs can naturally be more informative than paths and are demonstrated to handle complex logical formulas for reasoning \cite{TECHS,XinC24}. Unfortunately, efficiently retrieving valuable path information from the entire history to construct subgraphs and conducting reasoning in TKGs poses a significant challenge.

Deriving inspiration from the dual process theory \cite{evans1984heuristic,evans2003two,evans2008dual,sloman1996empirical}, a theory in cognitive science that explains the cognitive processes involving in thinking and reasoning.
Dual process theory argues that the reasoning system of humans first retrieves much relevant information for quick decisions through a shallow and intuitive process called System 1. Another controllable and logical reasoning process called System 2, performs deeper sequential thinking and complex reasoning based on System 1. Although existing extrapolation methods \cite{LiJGLGWC20,LiuZC0X022} explore the two-process theory to improve performance, they are still limited in terms of the interpretability of reasoning results.

In this paper, we propose a novel \textbf{Cogn}itive \textbf{T}emporal \textbf{K}nowledge \textbf{E}xtrapolation framework (CognTKE), which combines the
strengths of both embedding-based
and path-based methods.
CognTKE involves two processes: retrieving temporal relations from the TKG to build subgraphs and performing global shallow reasoning and local deep reasoning over subgraphs. Specifically,
we present a temporal cognitive relation directed graph (TCR-Digraph) as a core subgraph in CognTKE, which expands the scope of temporal relation paths to subgraphs, preserving crucial temporal relation paths for TKG reasoning. 
In TCR-Digraph, the first layer consists of query-related global one-hop historical facts and the subsequent layers consist of query-related local multi-hop historical facts.  
To effectively learn the historical temporal relation paths at different layers, we propose two reasoners sequentially encode global one-hop temporal relation and local multi-hop complex paths over TCR-Digraph, respectively. The two distinct learning processes bear resemblance to the concepts of System 1 and System 2 in the dual process theory, which sequentially performs fast thinking for intuitive reasoning and slow thinking for complex reasoning, respectively. 
In summary, our contributions are as follows:
\begin{itemize}
	\item  We propose a novel CognTKE framework for TKG extrapolation based on human cognition, which efficiently and sequentially handles the global shallow one-hop temporal relation reasoning and local deep complex multi-hop temporal path reasoning over TCR-Digraph.  
	\item We introduce a novel TCR-Digraph that captures crucial local and global temporal relation information based on TKG, providing more comprehensive evidence for the interpretation of TKG reasoning.
	\item Extensive experiments on four public datasets demonstrate that CognTKE achieves significant improvement in accuracy over the state-of-the-art baselines and has strong zero-shot reasoning ability.
\end{itemize}

\section{Related Work}
\subsubsection{Interpolation on TKGs} 
Interpolation on TKGs aims to infer the historical missing facts. Most interpolation works are developed by extending static KG reasoning methods.  For instance, TTransE \cite{TTransE} is a translation-based method and introduces time constraints between facts that have common entities. 
TNTComplEx\cite{TComplEX} is a tensor factorization method and proposes a 4th-order tensor factorization by extending time information into tensor factorization.
RotateQVS \cite{RotateQVS} models temporal evolutions as rotations in quaternion vector space, effectively capturing a variety of complex relational patterns. HGE-TNTComplEX \cite{HGE} projects temporal facts onto a product space of heterogeneous geometric subspaces and employs temporal-geometric attention mechanisms for interpolation.

\subsubsection{Extrapolation on TKGs} 
Extrapolation on TKGs aims to predict future unseen facts and can be classified as embedding-based methods and path-based methods.  
Some embedding-based methods explore the cyclic and repetitive patterns of historical facts from a global perspective. For example, CyGNet \cite{CyGNet} employs a copy-generation mechanism to learn query-relevant entities that are important in repeated historical facts.  
Other approaches study the local dependency of historical facts based on the local KG snapshots. For example, 
RE-GCN \cite{RE-GCN} models the evolution of entities and relations at each timestamp by using the local historical dependency.  CEN \cite{CEN} exploits a curriculum learning approach to model variable snapshot graph sequence lengths.
Recent embedding-based methods such as TiRGN \cite{TiRGN}, integrate the above two historical patterns to achieve more accurate reasoning. LogCL \cite{LogCL} adopts a contrastive learning  to enhance the representation of local and global representation of queries. 
Since the black box characteristics of embedding-based methods, it is difficult to provide explainable evidence for reasoning results.
Path-based methods generate temporal rules by sampling local temporal relations, and extend temporal rules to predict future facts. For example, 
TLogic \cite{Tlogic} mines the temporal logic rules from the loop relation path in a given time window and designs a strategy for evaluating the confidence of candidate temporal logic rules. TempValid \cite{Tempvalid} designs a rule-adversarial and a time-aware negative sampling strategies to obtain competitive results.
Our approach CognTKE is a combination of embedding-based and path-based methods, showcasing excellent explainability and zero-shot reasoning capabilities.
\begin{figure*}[htbp]
	\setlength{\abovecaptionskip}{0.5pt}
	\centering
	\includegraphics[width=0.9\linewidth]{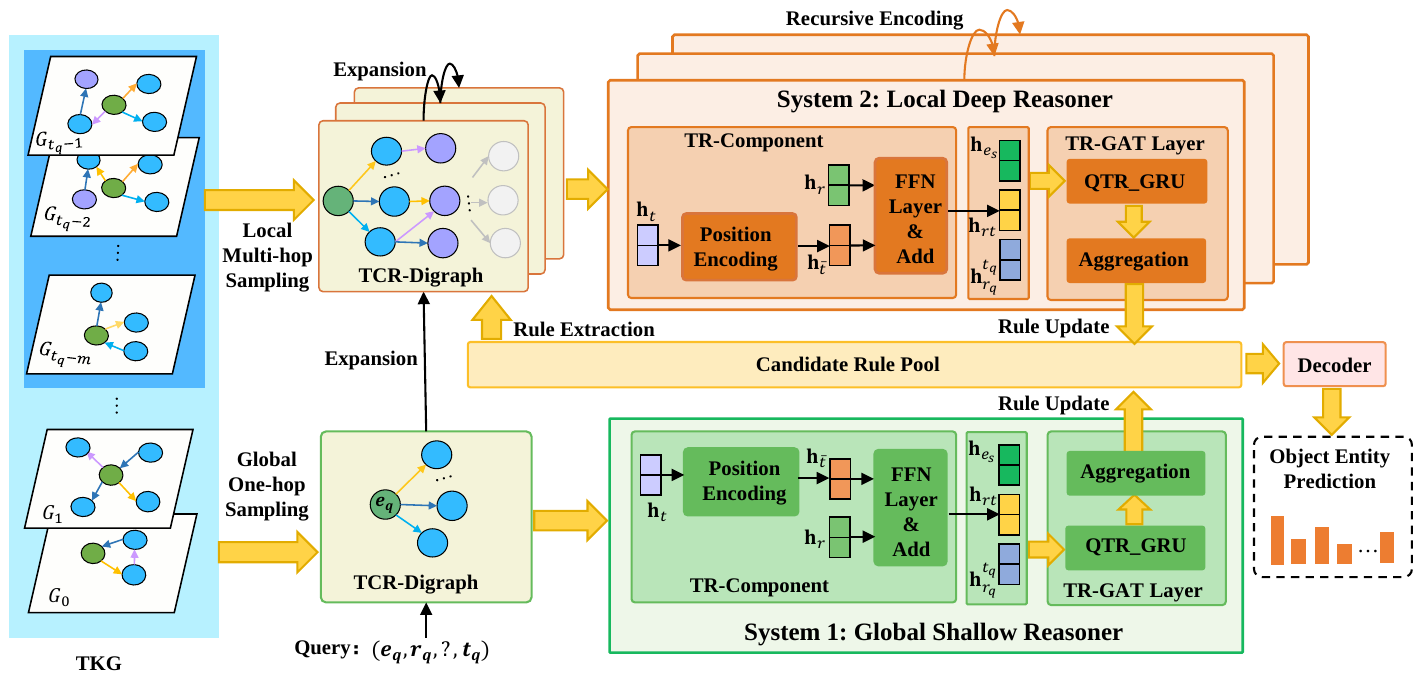}
	\caption{An illustrative diagram of the proposed CognTKE architecture.  }
	\label{fig:Main_Module}
\end{figure*}

\section{Our Approach}

\subsection{Preliminaries}
A TKG $\mathcal{G} $ is denoted as a sequence of KG snapshots, i.e.,  $\mathcal{G} = \{ \mathcal{G}_1, \mathcal{G}_2, ..., \mathcal{G}_{|\mathcal{T}|} \}$. The KG snapshot $\mathcal{G}_t = (\mathcal{E},\mathcal{R},\mathcal{F}_t)$ consists of a set of valid facts at $t \in \mathcal{T}$, where $\mathcal{E},\mathcal{R}, \mathcal{T}$ denote the set of entities,  relations, and timestamps, respectively; $\mathcal{F}_{t}$ denotes the set of facts at $t$. A fact (or an event) is represented in the
form of a quadruple $(e_s, r, e_o, t) \in \mathcal{F}_{t}$, where an edge $r$  serves as an association between the subject entity $e_{s}$ and object entity $e_{o}$ at $t$. 

Definition 1 (Temporal Relation Path over TKG). \textit{The temporal relation  path with length $l$  is a set of $l$ quadruples $(e_1, r_1, e_{2}, t_1) \rightarrow $ $\cdot\cdot\cdot \rightarrow (e_{l-1}, r_{l-1}, e_{l}, t_{l-1}) $ $\rightarrow(e_{l}, r_l, e_{l+1}, t_l)$  with $t_1 \leq$ $... \leq t_{l-1} \leq t_l$, that are connected origin entity $e_1$ to destination entity $e_{l+1}$ sequentially.}

Definition 2 (Problem Statement). \textit{TKG extrapolation aims to forecast the future unknown object entity (or subject entity) given a query $(e_{q}, r_{q}, ?,t_q)$ (or $(?, r_{q}, e_{q},t_q)$) according to previous historical KG snapshots $\{\mathcal{G}_{0}, \mathcal{G}_{1},..., \mathcal{G}_{t_{q}-1}\}$.}

Note that, the inverse relation quadruples $(e_s, r^{-1}, e_o,t)$ and identity quadruples $(e_s, r_{self}, e_s, t)$ are added to the TKG dataset, where $r_{self} \in \mathcal{R}$ is an identity relation. Without loss of generality, the TKG reasoning goal can be expressed as the prediction of object entities.

\subsection{Architecture Overview} 
The whole framework of CognTKE is shown in Figure \ref{fig:Main_Module}, involving the three processes: (1) Retrieving temporal relations from the TKG to build TCR-Digraph; (2) Performing global shallow encoding and local deep encoding over TCR-Digraph to generate crucial candidate rules to associate with the query;
(3) Based on the candidate rule pool, a decoder is employed to calculate the scores of entities.

\subsection{Temporal Cognitive Relation Digraph} 
Temporal relation paths have been proven to possess high transferability and interpretability in TKG \cite{Tlogic, XinC24}. However, the limitation lies in their ability to model complex multi-hop reasoning due to a small subset of local temporal relation paths learned in TKGs. Inspired by \cite{RED-GNN,WuW0WSL23,liang2024mines}, we introduce a temporal cognitive relation directed graph (TCR-Digraph) that provides rule-like explanatory and inductive capabilities to explore the significant historical chain of evidence. Next, we will introduce some definitions related to TCR-Digraph.

Definition 3 (Layered Graph\cite{layered}) \textit{The layered graph is a directed graph with exactly one source node and one sink node (destination node). All edges are directed, connecting nodes between consecutive layers and pointing from $L-1$-th layer to $L$-th layer. }

Definition 4 (Temporal Relation Directed Graph, TR-Digraph). \textit{The TR-Digraph $\bar{\mathcal{G}}_{e^{t_q}_{q},e^{t}_{o}|L}$ is a layered graph with the source entity (query entity) $e^{t_q}_{q}$ at query time $t_q$ and the sink entity (target entity) $e^{t}_{o}$ at history time $t$, where $t_q>t$ and $L$ is the number of layers in the TR-Digraph. Entities on the same layer are different from each other. Each edge $(r_{L}t_{L})^L$ formed by a combination of relation $r$ and history time $t$ is a direction, connecting an entity in the $L-1$-th layer to an entity in the $L$-th layer. $\bar{\mathcal{G}}_{e^{t_q}_{q},e^{t}_{o}|L}$ is set as $\emptyset$ if there is no temporal relation path connecting $e^{t_q}_{q}$ and $e^{t}_{o}$ with length $L$.}


Here, the quadruples with the reverse or identity relations are considered in the TR-Digraph. Through the above definition, any temporal relation path between query entity $e^{t_q}_{q}$ at query time $t_q$ and the target entity $e^{t}_{o}$ at specific history time $t$ in the TR-Digraph $\bar{\mathcal{G}}_{e^{t_q}_{q},e^{t}_{o}|L}$ is denoted as $e^{t_q}_{q} \rightarrow (r_1t_1)^1\rightarrow (r_2t_2)^2\rightarrow \cdot\cdot\cdot \rightarrow (r_Lt_L)^L \rightarrow e^{t}_{o}$ with length $L$ and $t_q \geq t_L$, where $(r_Lt_L)^{L}$ connects an entity in the $L-1$-layer to the entity in $L$-layer. 
From a high-level perspective, the TR-DiGraph can be seen as a subgraph composed of multiple temporal paths that originate from a single source node. This implies that we can construct the query TR-Digraph by iteratively retrieving the historical temporal relationship paths associated with the query entity $e^{t_q}_{q}$ in the TKG.
Thus, all temporal relation paths between the query entity $e^{t_q}_{q}$ and the target entity $e^{t}_{o}$ can be preserved in TR-Digraph. Unlike existing methods \cite{Tlogic,TITer} that impose temporal order constraints on consecutive temporal relations in each temporal relation path, we advocate for exploring additional features between temporal relations, such as causality, without limiting temporal order.

\begin{figure}[htbp]
	\setlength{\abovecaptionskip}{1pt}
	\centering
	\includegraphics[width=0.95\linewidth]{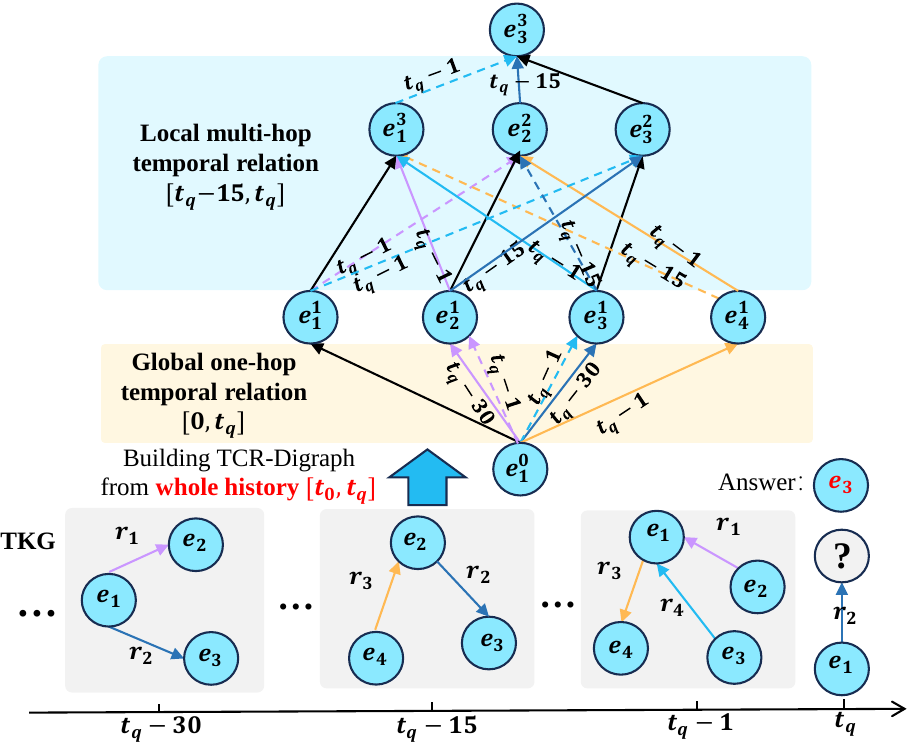}
	\caption{An illustrative of TCR-Digrph $\hat{\mathcal{G}}_{e_1,e_3|3}$ formed by the TKG.  }
	\label{fig:TCR-Digraph}
\end{figure}
In practice, such TR-Digraph can bring much earlier historical information that is not directly relevant to the query,  which not only makes the TR-Digraph construction and encoding expensive but also does not facilitate subsequent entity reasoning. Therefore, we propose an improved TCR-Digraph $\hat{\mathcal{G}}_{e^{t_q}_{q}, e^{t}_{o}|L}$ based on TR-Digraph, which contains important global one-hop history quadruples and recent local multi-hop history quadruples relevant to the query. 
To be specific, the first layer of the TCR-Digraph is built by retrieving the global query-related historical facts in the whole history time windows $[0,t_q]$. Based on candidate entities obtained in the previous layer, the subsequent layers of the
TCR-Digraph is built by further retrieving local historical facts associated with candidate entities in recent time windows $[t_q-m, t_q]$, where $m$ is the length of local time windows.
An example of TCR-Digraph $\hat{\mathcal{G}}_{e_1,e_3|3}$ in Figure \ref{fig:TCR-Digraph},
which is adopted to infer the new facts $(e_1, r_2, e_3, t_q)$ based on layered paths in TCR-Digraph $\hat{\mathcal{G}}_{e_1,e_3|3}$. 
For this purpose, TCR-Digraph preserves the crucial temporal relation paths associated with the query, making it more expressive and providing significant reasoning evidence and lower computational cost from the entire history.

\subsection{Recursive Encoding over TCR-Digraph}
As discussed above, the TCR-Digraph provides the interpretable chain of evidence consisting of global one-hop temporal relation and local multi-hop temporal relation paths. However, how to model different temporal relation paths on TCR-Digraph for temporal knowledge reasoning poses another problem. Inspired by the dual process theory of cognitive science \cite{evans2008dual}, we present a global shallow reasoner and a local deep reasoner consisting of  \textit{temporal relation component} (TR-Component) and \textit{temporal relation graph attention network} (TR-GAT).
The global shallow reasoner as the first decision process aims to carry out intuitive global one-hop reasoning, which can be seen as fast thinking in human cognition. The local deep reasoner as the second decision process performs local complex multi-hop reasoning, which can be regarded as slow thinking in human cognition. Although both reasoners perform different functions, similar pipelines of iteratively encoding each layer of the TCR-Digraph are employed to update the candidate rule pool. Note that each layer construction and encoding of TCR-Digraph is performed alternately. The candidate rule pool is a latent representation of candidate entities in TCR-Digraph, which is considered as working memory \cite{working} used to store updated rules for subsequent entity prediction.




\subsubsection{Temporal Relation Component}
The TR-Component aims to learn the representation of temporal relation in TCR-Digraph and serves as an input to TR-GAT. As historical facts associated with the query at different times have different effects on reasoning results, we adopt the positional encoding method \cite{transformer} to map relative time into embedding representations. Formally, the relative time encoding formula is as follows:
\begin{equation}
	\resizebox{0.68\linewidth}{!}{
		\begin{math}
			\begin{aligned}
				\mathbf{TE}^{(pos,2i)} = \sin(pos / 10000^{2i/d_{time}}),
			\end{aligned}
		\end{math}
	}
	\label{eq:1}
\end{equation}
\begin{equation}
	\resizebox{0.68\linewidth}{!}{
		\begin{math}
			\begin{aligned}
				\mathbf{TE}^{(pos,2i+1)} = \cos(pos / 10000^{2i/d_{time}}),
			\end{aligned}
		\end{math}
	}
	\label{eq:2}
\end{equation}
where $\mathbf{TE} \in \mathbb{R}^{|\mathcal{T}| \times d_{time}}$ is the embedding matrix of relative time, $pos$ denotes the location of the time value, $i$ denotes the dimension of the embedding presentation, and $d_{time}$ denotes the embedding size  of relative time, $\sin(\cdot)$ and $\cos(\cdot)$ are sine and cosine functions, respectively.

To incorporate the relative time information into the relation  while preserving the basic semantic information of the relation, we obtain the representation of the temporal relation $rt$ by the Feed-forward network (FFN) layer \cite{transformer} and add operation:
\begin{equation}
	\resizebox{0.85\linewidth}{!}{
		\begin{math}
			\begin{aligned}
				\mathbf{h}_{rt} = \sigma_1(\mathbf{W}_{2}(\sigma_1(\mathbf{W}_{1}[\mathbf{h}_r,\mathbf{v}_{\bar{t}}+\mathbf{b}_{1}])+\mathbf{b}_{2}) +  \mathbf{h}_r,
			\end{aligned}
		\end{math}
	}
	\label{eq:3}
\end{equation}
where $[,]$ denotes the concatenate operation of vectors, $\mathbf{h}_{r} \in \mathbf{H}$ denotes the embedding of relation, $\mathbf{H} \in \mathbb{R}^{ |\mathcal{R}| \times d}$ is an initialized relation matrix, $\mathbf{v}_{\bar{t}} \in \mathbf{TE}$ is the embedding of relative time, $\bar{t} = t_q-t$ is the relative time between the historical fact appearing time and the query time,  $\sigma_1$ is the $LeakyReLU$ activate function, $\mathbf{W}_{1} \in \mathbb{R}^{d\times d}, \mathbf{W}_{2} \in \mathbb{R}^{d\times (d+d_{time})}, \mathbf{b}_{1},  \mathbf{b}_{2}\in \mathbb{R}^{d}$ are all learnable parameters, where $d$ is the dimension size of the relation. Through the encoding of temporal relation, the quadruple $(e_s,r,e_o,t)$ in TKGs is converted into the form of triple $(e_s, rt,e_o)$.

\subsubsection{Temporal Relation Graph Attention Layer}
The goal of the TR-GAT layer is to propagate temporal relation information in TCR-Digraph $\hat{\mathcal{G}}_{e^{t_q}_{q}, e^{t}_{o} \mid L}$ centered on the query entity $e^{t_q}_{q}$ and update the candidate rule pool that is a latent representation of candidate answer entities. To preserve the ordering of  consecutive temporal relations in the TCR-Digraph during message propagation, we present a variant of GRU (QTR\_GRU) in TR-GAT to perform message
passing. The message passing function is as follows:
\begin{equation}
	\resizebox{0.7\linewidth}{!}{
		\begin{math}
			\begin{aligned}
				\mathbf{m}_{e_s,rt\mid r^{t_q}_{q}}^{l}=\operatorname{QTR\_GRU}(\mathbf{h}_{e_s}^{l-1}, \mathbf{h}_{rt}^{l}, \mathbf{h}_{r^{t_q}_{q}}^{l}),
			\end{aligned}
		\end{math}
	}
	\label{eq:5}
\end{equation}
where $\mathbf{h}_{r^{t_q}_{q}}^{l} \in\mathbb{R}^d$, $\mathbf{h}_{e_s}^{l} \in\mathbb{R}^d$ and $\mathbf{h}_{rt}^{l}\in\mathbb{R}^d$ are the representation of the query relation at query time $t_q$, the subject entity and temporal relation, respectively.  Note that, all entity representations are initialized to the zero vector before performing message passing. 


Since the construction of the TCR-Digraph $\hat{\mathcal{G}}_{e^{t_q}_{q},e^{t}_{o}|L}$ is dynamic and highly correlated with query entities and query time, while remaining independent of the query relation. Different queries with the same query entity at the same query time share the same TCR-Digraph, e.g., \textit{(The World Cup, Be\_held\_at, Qatar, 2022)}  and \textit{(The World Cup, Be\_won\_by, Argentina, 2022)},  leading to the same evidence being utilized for reasoning. 
Thus, we adopt the graph attention (GAT) \cite{GAT} mechanism to encode the query information into the attention weight to distinguish the importance of different edges in the TCR-Digraph.
The attention score $\mathbf{a}_{e_s,rt\mid r^{t_q}_{q}}^l$ for each triple $(e_s, rt, e_o)$ in the TCR-Digraph is calculated as follows: 
\begin{equation}
	\resizebox{0.92\linewidth}{!}{
		\begin{math}
			\begin{aligned}
				\mathbf{c}_{e_s,rt\mid r^{t_q}_{q}}^l=\sigma_2 (\mathbf{W}_6^l\sigma_1(\mathbf{W}_3^l \mathbf{h}_{e_s}^l+  \mathbf{W}_4^l \mathbf{h}_{rt}^{l}+ \mathbf{W}_5^l \mathbf{h}_{r^{t_q}_{q}}^{l})),
			\end{aligned}
		\end{math}
	}
	\label{eq:11}
\end{equation}
\begin{equation}
	\resizebox{0.72\linewidth}{!}{
		\begin{math}
			\begin{aligned}
				\mathbf{a}_{e_s,rt\mid r^{t_q}_{q}}^l=\frac{\exp(\mathbf{c}_{e_s,rt\mid r^{t_q}_{q}}^l)}{\sum_{(\tilde{e_s},\tilde{rt}) \in \mathcal{N}_{e_o}^l}\exp(\mathbf{c}_{\tilde{e_s}, \tilde{rt}\mid r^{t_q}_{q}}^l)},
			\end{aligned}
		\end{math}
	}
	\label{eq:12}
\end{equation}
where $\mathbf{W}^l_3, \mathbf{W}^l_4, \mathbf{W}^l_5, \mathbf{W}^l_6$ represent trainable weight matrices, $\mathbf{c}_{e_s,rt\mid r^{t_q}_{q}}^l$ denotes the attention weight, and $\mathcal{N}_{e_o}^l$ is the in-degree neighbors of the object entity $e_{o}$  in TCR-Digraph. Thus, the  candidate rule  representations of the aggregated entities are calculated as follows: 
\begin{equation}
	\resizebox{0.68\linewidth}{!}{
		\begin{math}
			\begin{aligned}
				\tilde{\mathbf{h}}_{e_o}^l=\mathbf{W}^{l}_7\sum_{(e_s, rt) \in \mathcal{N}_{e_o}^{l}}\mathbf{a}_{e_s,rt\mid r^{t_q}_q}^l \mathbf{m}_{e_s,rt\mid r^{t_q}_{q}}^l,
			\end{aligned}
		\end{math}
	}
	\label{eq:13}
\end{equation}
\begin{equation}
	\resizebox{0.4\linewidth}{!}{
		\begin{math}
			\begin{aligned}
				\hat{\mathbf{h}}_{e_o}^l=\frac{\tilde{\mathbf{h}}_{e_o}^l}{\sqrt{indegree(e_o)}}.
			\end{aligned}
		\end{math}
	}
	\label{eq:14}
\end{equation}

Among them,  $\mathbf{W}^{l}_{7}$ is the weight matrix. $indegree(e_o)$ denotes the number of in-degree object entities and serves as a scaling factor to adjust the entity representation.  
To further retain the feature of existing candidate entities, we utilize GRU \cite{GRU} to control the updating of representations of entities. Thus, the final representation of the candidate entity $e_o$ in $l$-th layer of TR-GAT can be calculated as follows:
\begin{equation}
	\resizebox{0.43\linewidth}{!}{
		\begin{math}
			\begin{aligned}
				\mathbf{h}_{e_o}^l = GRU (\mathbf{h}_{e_o}^{l-1},\hat{\mathbf{h}}_{e_o}^l). 
			\end{aligned}
		\end{math}
	}
	\label{eq:15}
\end{equation}
\section{Interpretable Reasoning and Analysis}
\subsubsection{Prediction}
After obtaining the final candidate entity representation that includes the appropriate temporal relation rule information, we adopt a simple MLP (Multilayer perceptron) as the decoder to calculate the prediction score of all entities as follows:
\begin{equation}
	\resizebox{0.72\linewidth}{!}{
		\begin{math}
			\begin{aligned}
				score(e_o)=f(e_{q}, r_{q}, e_o, t_q)=  \mathbf{W}_{o}\mathbf{h}_{e_o}^{l},
			\end{aligned}
		\end{math}
	}
	\label{eq:16}
\end{equation}
where $\mathbf{W}_{o}\in\mathbb{R}^{1\times d}$ is the weight matrix. 
If an entity is not included in the TCR-Digraph corresponding to the query, it will be assigned a score of 0.

We consider entity prediction as a multi-label classification task, and  utilize the multi-class log-loss to optimize the parameters $\Theta$ of CognTKE, which has been proven to be effective\cite{RED-GNN,CTD}, 
\begin{equation}
	\resizebox{0.75\linewidth}{!}{
		\begin{math}
			\begin{aligned}
				\mathcal{L} = \sum_{(e_{q}, r_{q}, e_o, t_q) \in \mathcal{F}_{train}}(-f(e_{q}, r_{q}, e_o, t_q)+ \\ log(\sum_{o^{\prime} \in \mathcal{E}}\exp({f(e_{q}, r_{q},e_{o^{\prime}}, t_q)})),
			\end{aligned}
		\end{math}
	}
	\label{eq:15}
\end{equation}
where $(e_{q}, r_{q}, e_o, t_q)$ denotes positive fact in training set $\mathcal{F}_{train}$, and $(e_{q}, r_{q}, e_{o^{\prime}}, t_q)$ denotes other fact with the same query $(e_{q}, r_{q},?,t_q)$. 

\subsubsection{Interpretable Analysis}
Although CognTKE acquires the graph structure of TCR-Digraph through the GAT mechanism, it still possesses the capacity to encode path-based logical rules of the same nature as those utilized in rule induction models, such as TLogic \cite{Tlogic} and TECHS\cite{TECHS}.

Theorem 1. \textit{Given a quadruple $(e_{q}, r_{q}, e_o, t_q)$, let $p$ be temporal relation $rt$, $\mathcal{C}$ be a set of
	temporal relation paths $e^{t_q}_{q} \rightarrow p^{1}_{k} \rightarrow p^{2}_{k} \rightarrow \cdots \rightarrow p^{L}_{k} \rightarrow e^{t}_{o}$( $t_q > t$) that are formed by a set of rules between $e^{t_q}_{q}$ and $e^{t}_{o}$ with the form:
	$$
	r_{q}(X, Y) \leftarrow p_{k}^1\left(X, Z_1\right) \wedge p_{k}^2\left(Z_1, Z_2\right) \wedge \cdots \wedge p_{k}^L\left(Z_{L-1}, Y\right),
	$$
	where $X, Y, Z_1, \ldots, Z_{L-1}$ are free variables that are bounded by unique entities. Assuming there exists a directed graph $\hat{\mathcal{G}}_{\mathcal{C}}$ that is built by $\mathcal{C}$, a parameter setting $\Theta$, and a threshold $\lambda \in(0,1)$ for CognTKE. $\hat{\mathcal{G}}_{\mathcal{C}}$ can equal to the TCR-digraph if CognTKE's edges have attention weights $\mathbf{c}_{e_s,rt\mid r^{t_q}_{q}}^{\ell}>\theta$ in $\hat{\mathcal{G}}_{e^{t_q}_{q}, e^{t}_{o} \mid L}$, where $\theta$ is a learned decision boundary parameter.}

According to Theorem 1, CognTKE is capable of encoding any temporal logical rule that corresponds to a path in the TCR-Digraph. This implies that if a set of temporal relation paths is highly correlated with the query quadruple, CognTKE can identify them through the attention weights, making them interpretable. 

\begin{table*}[t]
	\setlength\tabcolsep{1.8pt}
	\renewcommand{\arraystretch}{0.95}
	\centering
	\scalebox{0.85}{
		\begin{tabular}{cc|cccc|cccc|cccc|cccc}
			\toprule
			&\multirow{3}*{Model} 
			&\multicolumn{4}{c}{ICE14}&\multicolumn{4}{c}{ICE18}&\multicolumn{4}{c}{ICE05-15}&\multicolumn{4}{c}{WIKI} \\
			\cmidrule(lr){3-6} \cmidrule(lr){7-10} \cmidrule(lr){11-14} \cmidrule(lr){15-18}
			
			&&MRR &Hits@1 &Hits@3 &Hits@10 &MRR &Hits@1 &Hits@3 &Hits@10 &MRR &Hits@1 &Hits@3 &Hits@10 &MRR &Hits@1 &Hits@3 &Hits@10\\
			\midrule
			&CyGNet & 35.05  & 25.73  & 39.01  & 53.55  & 24.93  & 15.90  & 28.28  & 42.61  & 36.81  & 26.61  & 41.63  & 56.22  & 33.89  & 29.06  & 36.10  & 41.86  \\
			&TANGO & 36.80  & 27.43  & 40.90  & 54.94  & 28.68  & 19.35  & 32.17  & 47.04  & 42.86  & 32.72  & 48.14  & 62.34  & 50.43  & 48.52  & 51.47  & 53.58  \\
			&RE-GCN & 40.39  & 30.66  & 44.96  & 59.21  & 30.58  & 21.01  & 34.34  & 48.75  & 48.03  & 37.33  & 53.85  & 68.27  & 77.55  & 73.75  & 80.38  & 83.68  \\
			&CEN & 42.20  & 32.08  & 47.46  & 61.31  & 31.50  & 21.70  & 35.44  & 50.59  & 46.84  & 36.38  & 52.45  & 67.01  & 78.05  & 75.05  & 81.90  & 84.90  \\
			&TITer & 40.87  & 32.28  & 45.45  & 57.10  & 29.98  & 22.05  & 33.46  & 44.83  & 47.69  & 37.95  & 52.92  & 65.81  & 75.50  & 72.96  & 77.49  & 79.02  \\
			&TLogic & 43.04  & 33.56  & 48.27  & 61.23  & 29.82  & 20.54  & 33.95  & 48.53  & 46.97  & 36.21  & 53.13  & 67.43  &   -    &  -     & -      & - \\
			&TiRGN &44.04  & 33.83  & 48.95  & 63.84  & \underline{33.66}  & 23.19  & 37.09  & \underline{54.22}  & 50.04  & 39.25  & 56.13 & \underline{70.71}  & \underline{81.65}  & \underline{77.77}  & 85.12  & 87.08  \\
			&CENET & 39.02 & 29.62 & 43.23 & 57.49 & 27.85 & 18.15 & 31.63 & 46.98 & 41.95 & 32.17 & 46.93 & 60.43 & 40.52 & 32.91 & 45.11 & 53.08 \\
			&RETIA & 42.76  & 32.28  & 47.77  & 62.75  & 32.43  & 22.23  & 36.48  & 52.94  & 47.26  & 36.64  & 52.90  & 67.76  & 78.58  & 74.79  & 81.45  & 84.60  \\
            &TECHS & 43.88  & 34.59  & 49.36  & 61.95  & 30.85  & 21.81  & 35.39  & 49.82  & 48.38  & 38.34  & 54.69  & 68.92  & 75.98  & -  & -  & 82.89  \\
            &TempValid & \underline{45.78}  & \underline{35.50}  & \textbf{51.34}  & \textbf{65.06}  & 33.50  & \underline{23.91}  & \underline{37.89}  & 52.33  & \underline{50.31}  & \underline{39.46}  & \underline{56.71}  & 70.55  & 83.19  & 74.64  & \textbf{90.12}  & \textbf{97.54}  \\
			\midrule
			&CognTKE &\textbf{46.06}  & \textbf{36.49}  & \underline{51.11}  & \underline{64.49}  & \textbf{35.24}  & \textbf{25.21}  & \textbf{39.93}  & \textbf{54.71}  & \textbf{53.13}  & \textbf{42.62}  & \textbf{59.42}  & \textbf{72.70}  & \textbf{83.21}  & \textbf{80.01}  & \underline{86.07}  & \underline{87.34}  \\
			\bottomrule
		\end{tabular}
	}
    \label{table:over-results}
	\setlength{\abovecaptionskip}{3pt}
	\caption{The prediction results of MRR and Hits@1/3/10 on all datasets.}
\end{table*}	

\section{Experiments}
\subsubsection{Datasets} To evaluate CognTKE on entity prediction task, we adopt four benchmark datasets that are widely used for TKG extrapolation,  including  ICE14, ICE18, ICE05-15 \cite{RE-NET}, and WIKI \cite{RE-GCN}. 
Following the preprocessing strategy \cite{RE-GCN,TANGO, CEN}, we split all datasets into training, validation, and test sets with the proportions of 80\%/10\%/10\% based on chronological order. 

\subsubsection{Baselines}
To demonstrate the effectiveness of CognTKE for TKG reasoning, CognTKE is compared with up-to-date TKG extrapolation methods including CyGNet \cite{CyGNet},
TANGO \cite{TANGO}, TITer \cite{TITer},  RE-GCN \cite{RE-GCN}, CEN \cite{CEN}, TiRGN \cite{TiRGN}, TLogic \cite{Tlogic}, CENET \cite{CENET}, RETIA \cite{RETIA}, TECHS \cite{TECHS}, and TempValid\cite{Tempvalid}.

\subsubsection{Evaluation Metrics}
We utilize two evaluation metrics: mean reciprocal rank (MRR) and Hits ($k$=1, 3, 10), which are commonly employed to assess the effectiveness of TKG reasoning methods. 
We follow works \cite{TANGO,TITer} to report the experimental results with the time-aware filtered setting, which only filters out the quadruples occurring at the query time.  Note that, all experimental results in this paper regarding MRR and Hits@1/3/10 are reported as percentages.


%

\begin{table}[htbp]
	\setlength\tabcolsep{3.5pt}
	\renewcommand{\arraystretch}{0.85}
	\centering
    \label{table:ablation}
	\scalebox{0.85}{
		\begin{tabular}{c|cc|cc|cc}
			\toprule
			\multirow{2}{*}{Model}
			&\multicolumn{2}{c}{ICE14}&\multicolumn{2}{c}{ICE18}&\multicolumn{2}{c}{ICE05-15}\\
			\cmidrule(lr){2-3} \cmidrule(lr){4-5} \cmidrule(lr){6-7} 
			&MRR  &Hits@3 &MRR &Hits@3 &MRR  &Hits@3  \\
			\midrule
			CognTKE & \textbf{46.06} &  \textbf{51.11} &  \textbf{35.24} &  \textbf{39.93} &  \textbf{53.13} &  \textbf{59.42} \\
			\textit{-w/o System1}  & 36.12   & 40.22    & 29.01    & 32.58    & 40.61    & 45.08   \\
			\textit{-w/o System2} &44.88  & 50.08   & 33.45    & 38.01    & 44.49   & 50.16    \\
			\textit{-w/o Ti}  &45.44    & 50.53   & 34.01   & 38.47  & 51.49   & 57.85   \\
			\textit{-w/o QTR}  &44.72   & 49.57  & 34.55    & 39.12    & 49.82   & 56.03 \\
			\bottomrule
		\end{tabular}
	}
	\setlength{\abovecaptionskip}{3pt}
	\caption{The results of ablation study on ICE14, ICE18 and ICE05-15 datasets.}
\end{table}

\begin{table*}[htbp]
	\setlength\tabcolsep{4pt}
	\renewcommand{\arraystretch}{0.8}
	\centering
	\caption{The zero-shot reasoning results on ICE14, ICE18 and ICE05-15 datasets.}
	\scalebox{0.85}{
		\begin{tabular}{c|cccc|cccc|cccc}
			\toprule
			Model & \multicolumn{3}{c|}{CognTKE} & TempValid & \multicolumn{3}{c|}{CognTKE} & TempValid & \multicolumn{3}{c|}{CognTKE} & \multicolumn{1}{c}{TempValid} \\
			\midrule
			Test  & \multicolumn{4}{c|}{ICE14}  & \multicolumn{4}{c|}{ICE18}  & \multicolumn{4}{c}{ICE05-15} \\
			\midrule
			Train & ICE14 & ICE18 & ICE05-15 & ICE14 & ICE14 & ICE18 & ICE05-15 & ICE18 & ICE14 & ICE18 & ICE05-15 & ICE05-15 \\
			\midrule
                MRR   & 46.06 & 46.01 & \textbf{46.71} & 45.78 & 33.07 & \textbf{35.24} & 34.21 & 33.50 & 49.99 & 49.86 & \textbf{53.13} & 50.31 \\
                Hits@1 & 36.49 & 36.31 & \textbf{37.11} & 35.50 & 23.3  & \textbf{25.21} & 24.27 & 23.91 & 39.86 & 39.52 & \textbf{42.62} & 39.46  \\
                Hits@3 & 51.11 & 51.41 & \textbf{51.77} & 51.34 & 37.38 & \textbf{39.93} & 38.75 & 37.89 & 56.03 & 55.96 & \textbf{59.42} & 56.71  \\
			\bottomrule
		\end{tabular}
	}
	\setlength{\abovecaptionskip}{3pt}
    \label{table:zero-shot}
\end{table*}

\subsection{Overall Results}
The overall experimental results of CognTKE and baselines on four benchmark datasets are displayed in Table 1. The best results are marked in bold, and the second-best results are reported using underlining. 

It can be seen that CognTKE consistently outperforms the best path-based baseline TempValid on the ICE18 and ICE05-15 datasets. On ICE14 and WIKI datasets, CognTKE achieves better results compared to TempValid on MRR and Hits@1 metrics, and worse on Hits@3/10 metrics. Since TempValid's cyclic temporal rules can effectively model simple temporal rules on the ICE14 and WIKI datasets, it struggles to mine more complex temporal rules on the more challenging ICE18 and ICE05-15 datasets. 
This demonstrates that our CognTKE is capable of modeling complex temporal relation rules.
Compared with other path-based methods TLogic, TlTer and TECHS, CognTKE achieves significant performance improvement. This is mainly because the path evidence obtained by the TLogic, TlTer and TECHS is restricted, while the TCR-Digraph of CognTKE can enjoy more temporal relation rule semantics.
 
The embedding method TiRGN outperforms other embedding methods because TiRGN integrates local and global historical information, but the interpretability of reasoning results is poor. By integrating the global one-hop temporal relations and the local multi-hop paths,  CognTKE not only outperforms the performance of TiRGN, but also has strong interpretability.  

\subsection{Ablation Study}
To further better understand each component of CognTKE that contributes to the prediction results, we conduct ablation studies on ICE14, ICE18, and ICE05-15 datasets.  As reported in Table 2. \textit{-w/o System1} denotes a variant of CognTKE that only considers the local deep reasoner; \textit{-w/o System2} denotes a variant of CognTKE that only uses the global shallow reasoner; \textit{-w/o Ti} denotes a variant of CognTKE that doesn't consider the TR-Component; \textit{-w/o QTR} denotes a variant of CognTKE that doesn't use the QTR\_GRU for message aggregation, but instead uses addition.

From the results in Table 2, it can be observed that  \textit{-w/o System1} and \textit{-w/o System2} perform consistently worse than CognTKE on ICE14, ICE18, and ICE05-15 datasets. This demonstrates that the removal of either the global shallow reasoner or the local deep reasoner results in a decline in performance.
Another interesting observation is that the performance of \textit{-w/o System1} is inferior to \textit{-w/o System2}. The reason for this phenomenon is that 
\textit{-w/o System2} can directly answer the majority of query questions by the global shallow reasoner, since most query questions are associated with global one-hop historical facts that appeared in history. On the other hand, \textit{-w/o System1}  primarily focuses on local complex multi-hop reasoning, and many query problems are unrelated to local historical facts. So \textit{-w/o System1} is difficult to infer the query problems associated with longer historical facts. Hence, combining the global shallow reasoner and the local deep reasoner leads to more accurate prediction results.

The results of \textit{-w/o Ti} denote that the TR-Component is useful, since the TR-Component can help CognTKE distinguish time intervals between temporal relations and queries.  The performance of \textit{-w/o QTR} denotes the  QTR\_GRU is beneficial for the CognTKE to aggregate and update messages in TCR-Digraph. The main reason is that  QTR\_GRU considers the order between adjacent temporal relations, and such an order will guide CognTKE to infer more accurate results.
\subsection{Zero-Shot Reasoning}
The proposed CognTKE is an inductive approach that allows for transferability to new datasets sharing the same relations as the training dataset, achieving unknown entity prediction through zero-shot reasoning. To evaluate the effectiveness of zero-shot reasoning in CognTKE, we conduct experiments on ICEWS series datasets: ICE14, ICE18, and ICE05-15 datasets. ICEWS series datasets share a majority of the same relations. The results are shown in Table 3, we also bold the best results. It can be seen that CognTKE still performs better than the best path-based baseline TempValid, when ICE14 and ICE18 are used as test sets. This demonstrates that CognTKE exhibits strong capabilities for handling new data migrations. When ICE05-15 is considered as the test set, the prediction results trained on ICE14 and ICE18 datasets are close to TempValid, but worse than that trained on ICE05-15 dataset. The main reason is that the time span of ICE14 and ICE18 datasets is smaller than the span of ICE05-15, which cannot provide enough historical facts to generate corresponding temporal relation rules.
\begin{figure}[htbp]
	\setlength{\abovecaptionskip}{0.5pt}
	\centering
	\includegraphics[width=0.9\linewidth]{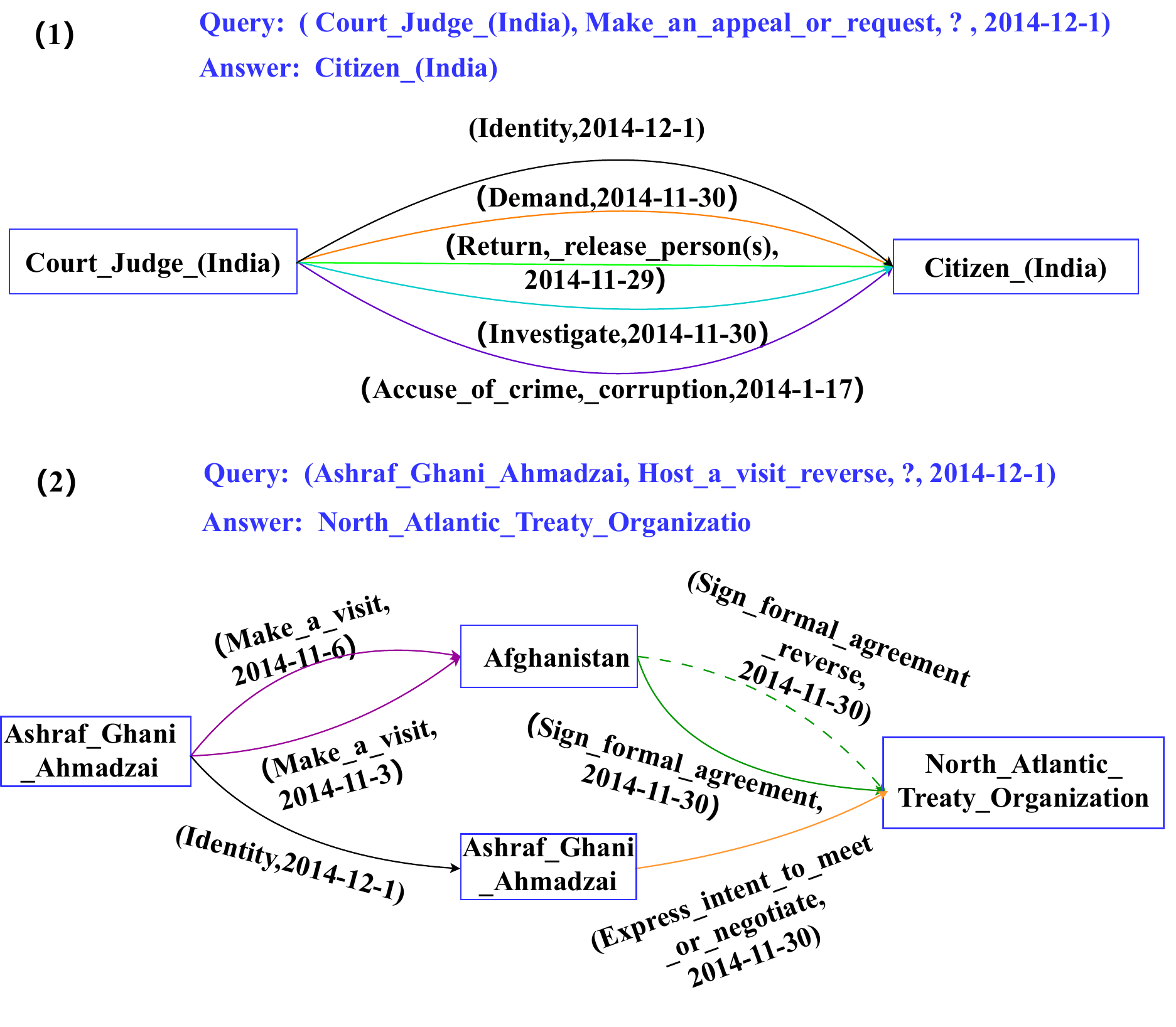}
	\caption{Visualization of the learned structures. Dashed lines mean inverse relations. entities are indicated by the blue rectangles.
	}
	\label{fig:Case-Study}
\end{figure}
\subsection{Case Study}
We visualize two cases of TCR-digraphs learned by CognTKE on ICE14 dataset. 
We eliminate edges in the TCR-Digraph whose attention weights are below 0.4.
Two queries with the learned structure consisting of temporal relations are shown in Figure \ref{fig:Case-Study}.  
For the first query, we can observe that CognTKE can directly obtain the answer \textit{Citizen\_(India)} from the global one-hop temporal relation, without further local multi-hop judgment. For the second query, although the quadruple \textit{(Ashraf\_Ghani\_Ahmadzai, Express\_intent\_to\_meet\_or\_negotiate, North\_Atlantic\_Treaty
	\_Organization, 2014-11-30) }  directly contains the answer entity, it is still necessary to traverse the identity quadruple to reach the answer entity. The reason for this is mainly attributed to the lower degree of correlation between the two relations \textit{Host\_a\_visit\_reverse } and \textit{Express\_intent\_to\_meet
	\_or\_negotiate}. The history relation \textit{Make\_a\_visit}  is semantically similar to the query relation, so it gets more weight. However, as the answer entity is not found in one-hop quadruples, exploring local two-hop quadruples becomes essential to infer a more accurate answer.
Two cases in Figure \ref{fig:Case-Study} further demonstrate that CognTKE has a strong interpretive capability. 

\section{Conclusion}
In this paper, we propose a novel cognitive temporal knowledge reasoning framework (CognTKE) according to dual process theory. In CognTKE, we introduce a novel TCR-Digraph that
consists of significant local and global historical temporal relation paths associated with the query. 
A global shallow reasoner and a local deep reasoner are designed to perform inductive reasoning over the TCR-Digraph. 
TR-Component and TR-GAT in both reasoners are employed to distinguish the temporal relation paths in TCR-Digraph most relevant to the query.
Extensive experiments on four datasets demonstrate that CognTKE outperforms state-of-the-art baselines and exhibits excellent zero-shot reasoning capabilities and strong interpretability.

\section{Acknowledgments}
This work is supported by the National Natural Science Foundation of China under Grant 62406022.

\bibliography{aaai25}

\newpage

\section{Appendix}
\subsection{A Some Details in Our Approach}
\subsubsection{A.1 Structure Diagram of QTR\_GRU. } 
\begin{figure}[htbp]
	\centering
	\includegraphics[width=0.9\linewidth]{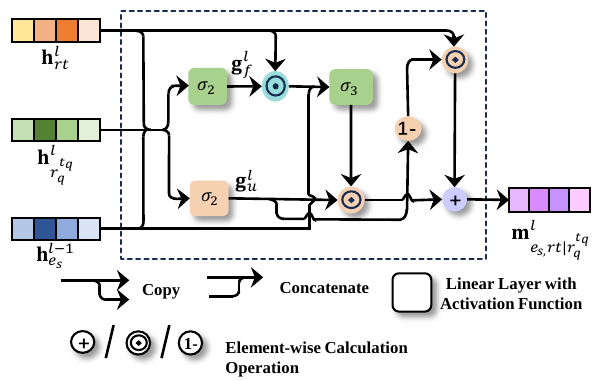}
	\caption{The Component of QTR\_GRU.
	}
	\label{fig:QTR}
\end{figure}

The QTR\_GRU component in CognTKE is shown in Figure \ref{fig:QTR}, the detailed calculation process of  QTR-GRU can be described as:
\begin{equation}
	\resizebox{0.65\linewidth}{!}{
		\begin{math}
			\begin{aligned}
				\mathbf{g}_{u}^{l}= \sigma_2(\mathbf{W}^{l}_u[\mathbf{h}_{e_s}^{l-1}, \mathbf{h}_{rt}^{l}, \mathbf{h}_{r^{t_q}_{q}}^{l}]+\mathbf{b}^{l}_u), 
			\end{aligned}
		\end{math}
	}
	\label{eq:6}
\end{equation}
\begin{equation}
	\resizebox{0.65\linewidth}{!}{
		\begin{math}
			\begin{aligned}
				\mathbf{g}_{f}^{l} = \sigma_2(\mathbf{W}^{l}_f[\mathbf{h}_{e_s}^{l-1}, \mathbf{h}_{rt}^{l}, \mathbf{h}_{r^{t_q}_{q}}^{l}]+\mathbf{b}^{l}_f),
			\end{aligned}
		\end{math}
	}
	\label{eq:7}
\end{equation}
\begin{equation}
	\resizebox{0.69\linewidth}{!}{
		\begin{math}
			\begin{aligned}
				\mathbf{h}_{c}^{l}=\sigma_3( \mathbf{W}^{l}_c(\mathbf{h}_{rt}^{l}+(\mathbf{g}_{f}^{l}\odot \mathbf{h}_{e_s}^{l-1}))+\mathbf{b}^{l}_c), 
			\end{aligned}
		\end{math}
	}
	\label{eq:8}
\end{equation}
\begin{equation}
	\resizebox{0.7\linewidth}{!}{
		\begin{math}
			\begin{aligned}
				\mathbf{m}_{e_s,rt\mid r^{t_q}_{q}}^{l}=(1-\mathbf{g}_{u}^l)\odot \mathbf{h}_{s}^{l-1}+\mathbf{g}_{u}^{l}\odot \mathbf{h}_{c}^{l}. 
			\end{aligned}
		\end{math}
	}
	\label{eq:9}
\end{equation}

Among them, $\odot$ represents the Hadamard product, $\mathbf{g}_f$, $\mathbf{g}_u$ and $\mathbf{h}_c$ denote the forget gate, the update gate, and the hidden state in QTR\_GRU, respectively;  $\mathbf{W}^{l}_u, \mathbf{W}^{l}_f\in\mathbb{R}^{d\times 3d}, \mathbf{W}^{l}_c\in\mathbb{R}^{d\times d}, \mathbf{b}^{l}_u, \mathbf{b}^{l}_f, \mathbf{b}^{l}_c \in\mathbb{R}^d$ all denote trainable parameters. $\sigma_2$ and $\sigma_3$ represent the \textit{sigmoid} activate function and \textit{tanh} activate function, respectively. 

\subsubsection{A.2 Pseudocode. }
The algorithm for the TCR-Digraph encoding in CognTKE is described in Algorithm 1.

\begin{algorithm}[tb]  
	\caption{The encoder of CognTKE }  
	\label{Algorithm 1}   
		\textbf{Input}: Query $(e_{q}, r_{q},?, t_q)$, a KG snapshot sequence $\{ \mathcal{G}_{1}, \mathcal{G}_{2}, ..., \mathcal{G}_{t_q-1} \}$.\\
		\textbf{Output}: Prediction scores for all queries.
        \begin{algorithmic}[1] 
		\STATE Initial the embeddings of relations and times, query entity embedding $\mathbf{h}_{e_q^{t_q}}^1=0$, entity set  $\hat{\mathcal{E}}^1=\{{e_{q}^{t_q}}\}$ in the TCR-Digraph, quadruple set $\mathcal{D}^1=\emptyset$, $i=1$.
		\WHILE{$t< |\mathcal{T}| $}
		\STATE Sample neighbor quadruples based on $\hat{\mathcal{E}}^i$ from the KG snapshot sequence $\{ \mathcal{G}_{1}, \mathcal{G}_{2}, ..., \mathcal{G}_{t_q-1} \}$ and expand the TCR-Digraph. 
		\STATE $i+=1$.
		\STATE $\mathcal{D}^i=\{(e_{s}, r, e_{o}, t) | e_{s} \in \mathcal{E}^{i-1} \}$,
		$\mathcal{E}^i=\{e_{o} | e_{s} \in \mathcal{E}^{i-1}\wedge(e_{s}, r, e_{o}, t) \in \mathcal{F}_{global} \}\cup \mathcal{E}^{i-1}$. 
		\STATE Obtain temporal relation representation $\mathbf{h}_{rt}$ by Eq. (1)-(3).
		\STATE Update $\mathbf{h}_{e_o}$ by Eq.(4)-(9).
		\FOR{$i<=L$}  
		\STATE Sample neighbor quadruples based on $\hat{\mathcal{E}}^i$ from the KG snapshot sequence $\{ \mathcal{G}_{t_q-m}, ..., \mathcal{G}_{t_q-1} \}$ and expand the TCR-Digraph.
		\STATE $i+=1$.
		\STATE $\mathcal{D}^i=\{(e_{s}, r, e_{o}, t) | e_{s} \in \mathcal{E}^{i-1}\wedge(e_{s}, r, e_{o}, t) \in \mathcal{F}_{local} \}$,
		$\mathcal{E}^i=\{e_{o} | e_{s} \in \mathcal{E}^{i-1}\wedge(e_{s}, r, e_{o}, t) \in \mathcal{F}_{local} \}\cup \mathcal{E}^{i-1}$.
		\STATE Obtain temporal relation representation $\mathbf{h}_{rt}$ by Eq. (1)-(3).
		\STATE Update $\mathbf{h}_{e_o}$ by Eq.(4)-(9).
		\ENDFOR
		\STATE Calculate the prediction scores by Eq.(10).
		\ENDWHILE
	\end{algorithmic}  
\end{algorithm} 

\begin{figure}[htbp]
	\setlength{\abovecaptionskip}{0.5pt}
	\centering
	\subfigure[ICE14]{
		\begin{minipage}[t]{0.48\linewidth}
			\centering
			\includegraphics[width=1\linewidth]{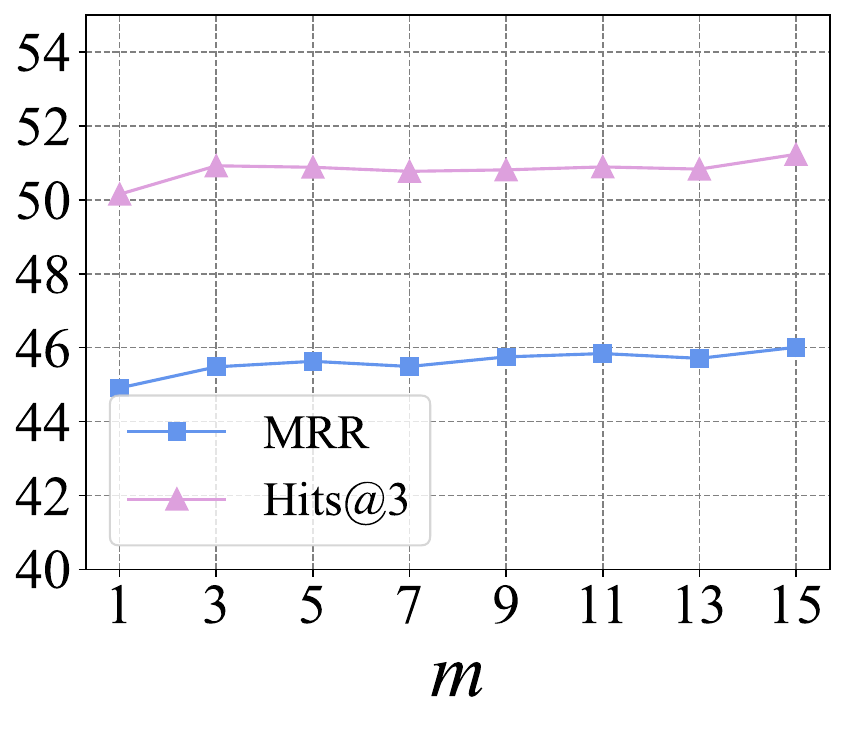}
		\end{minipage}
	}%
	\subfigure[ICE05-15]{
		\begin{minipage}[t]{0.48\linewidth}
			\centering
			\includegraphics[width=1\linewidth]{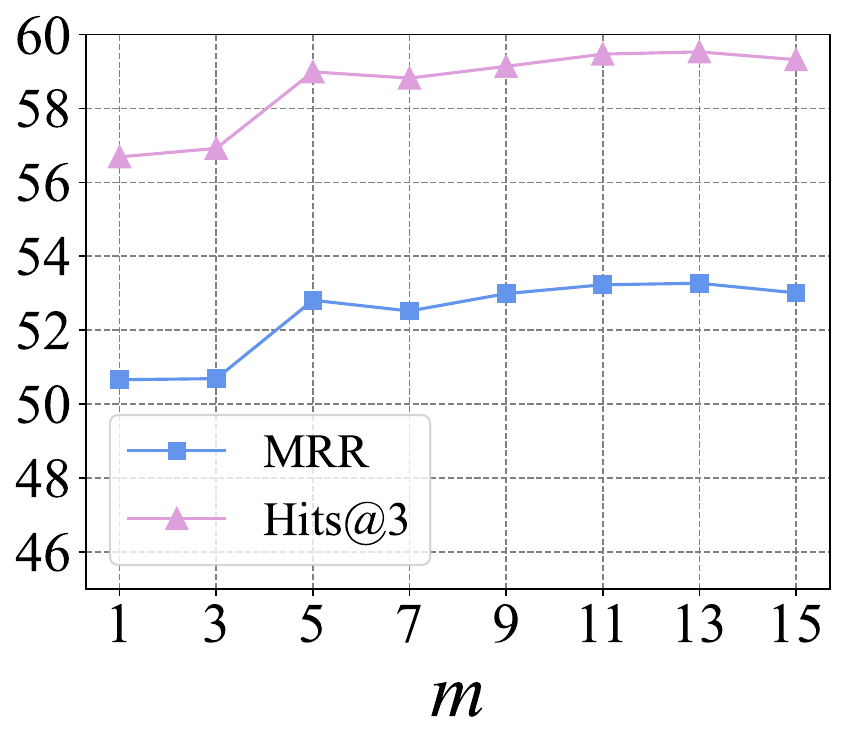}
		\end{minipage}
	}
	\caption{Study on the length of the local time window $m$ on ICE14 and ICE05-15 datasets.}
	\label{fig:length}
\end{figure}

\begin{figure}[htbp]
	\setlength{\abovecaptionskip}{0.5pt}
	\centering
	\subfigure[ICE14]{
		\begin{minipage}[t]{0.48\linewidth}
			\centering
			\includegraphics[width=1\linewidth]{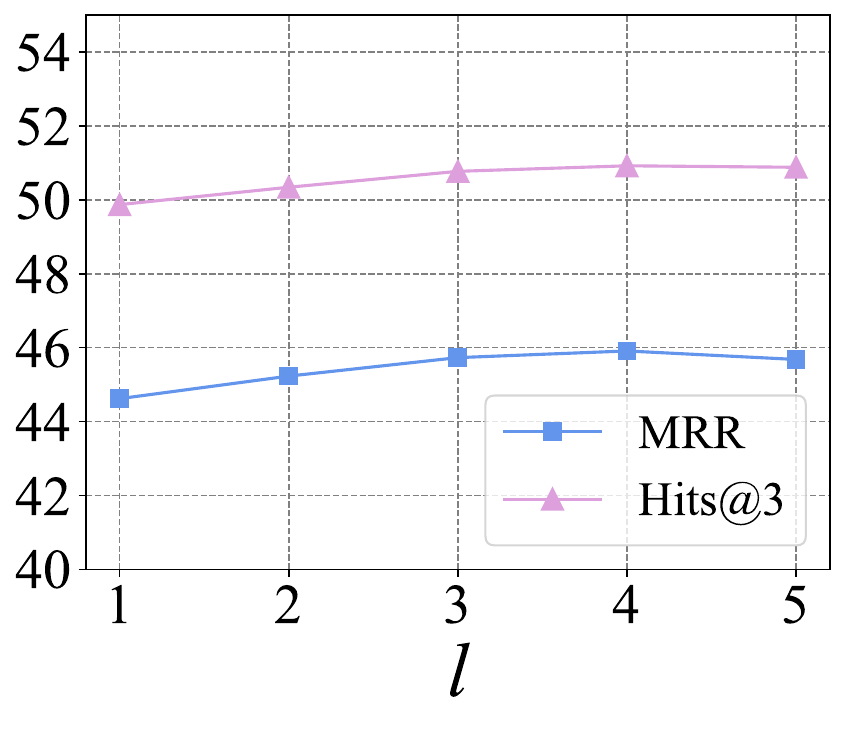}
		\end{minipage}
	}%
	\subfigure[ICE05-15]{
		\begin{minipage}[t]{0.48\linewidth}
			\centering
			\includegraphics[width=1\linewidth]{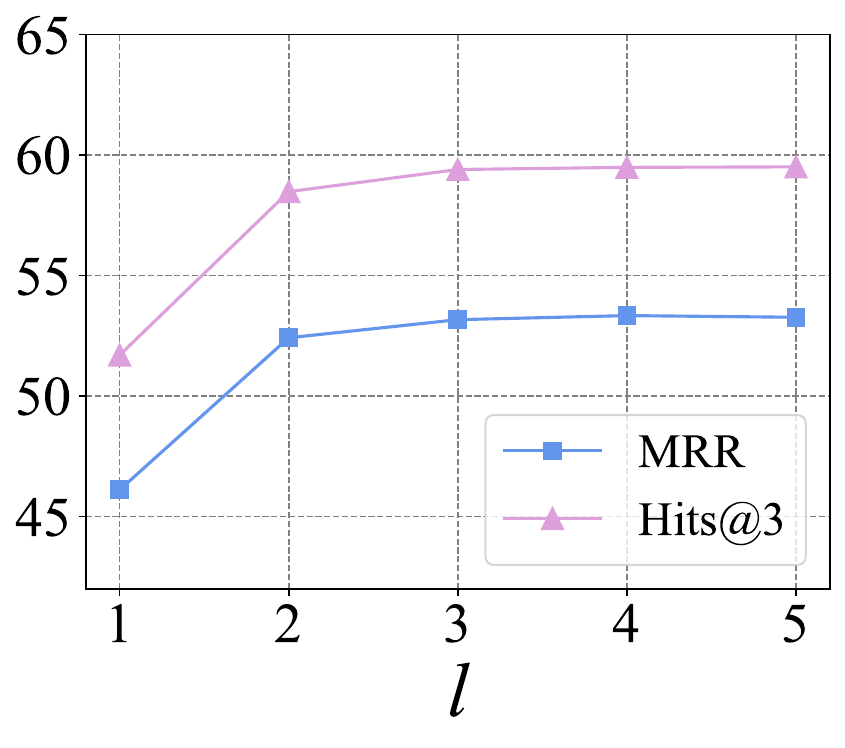}
		\end{minipage}
	}
	\caption{Study on the layers of TR-GAT $l$ on ICE14 and ICE05-15 datasets.}
	\label{fig:layers}
\end{figure}

\subsection{B Experiment Details}
\subsubsection{B.1 Dataset Statistics.} Table 4 shows the statistics of the four datasets ICE14, ICE18, ICE05-15, and WIKI.

\begin{table}[h]
	\setlength\tabcolsep{6pt}
	\renewcommand{\arraystretch}{1}
	\centering
	\scalebox{0.8}{
		\begin{tabular}{c|cccc}
			\toprule
			Dataset &ICE14  &ICE18 &ICE05-15 & WIKI     \\
			\midrule
			Entities &7,128  &23,033 &10,094 & 12,554  \\
			Relations & 230 & 256& 251 & 24\\
			Training & 74,845 & 373,018& 368,868& 539,286\\
			Validation & 8,514  &45,995 &46,302 &67,538\\
			Test & 7,371 &49,545 &46,159 &63,110\\
			Time granularity & 24 hours &24 hours &24 hours & 1 year\\
			Snapshot numbers & 365 & 365 & 4017 & 232   \\
			\bottomrule 
		\end{tabular}
		\label{table:dataset}
	}
	\setlength{\abovecaptionskip}{3pt}
	\caption{Details of the TKG datasets}
\end{table}

\subsubsection{B.2 Implementation Details.} For all datasets, the embedding size $d$ is set to 64, the time embedding size is set to 32,  the learning rate is set to 0.001, the batch size is set to 128, the length of local time windows $m$ is set to 15, the number of layers in TR-GAT is set to 4. CognTKE is implemented using PyTorch and PyG.  

The CognTKE employs Adam to optimize the parameters of the model. To train the CognTKE, we utilize an NVIDIA Tesla A100 GPU for 20 epochs in a Linux machine and a mixed precision training way.

For extrapolation baselines, their results with the time-aware filter setting are reported under the default parameters. For fairness of comparison, results of CEN and RETIA  are reported under the offline setting that is adopted to other baselines.

\subsection{Parameters Analysis}
We conduct experiments on ICE14 and ICE05-15 datasets to further analyze the impact of parameters in CognTKE, including the length of local time windows $m$, the number of layers of TR-GAT $l$, the embedding size of relation $d$, and the embedding size of time $d_{time}$. 

From the results in Figure \ref{fig:length}, it can be seen that the performance increases slowly with the size of the time window on ICEW14 and ICEW05-15 datasets since larger time windows can learn more local temporal relation rules. However, setting the time window too large will lead to more computational overhead. Selecting an appropriate time window is crucial to strike a balance between performance and computational overhead.

Since CognTKE is an inductive method, the number of layers in TR-GAT can be viewed as the length of the reasoning path. From the results in Figure \ref{fig:layers}, it is observed that as the number of layers increases, the performance improves slowly and then levels off on ICE14 and ICE05-15 datasets. The reason is that the local KG snapshot is small in scale and the historical information of KG snapshots are basically retrieved by using 4-hop. 
Therefore, the performance of CognTKE ceases to improve when the number of layers in TR-GAT exceeds 4.

\begin{figure}[htbp]
	\setlength{\abovecaptionskip}{0.5pt}
	\centering
	\subfigure[ICE14]{
		\begin{minipage}[t]{0.48\linewidth}
			\centering
			\includegraphics[width=1\linewidth]{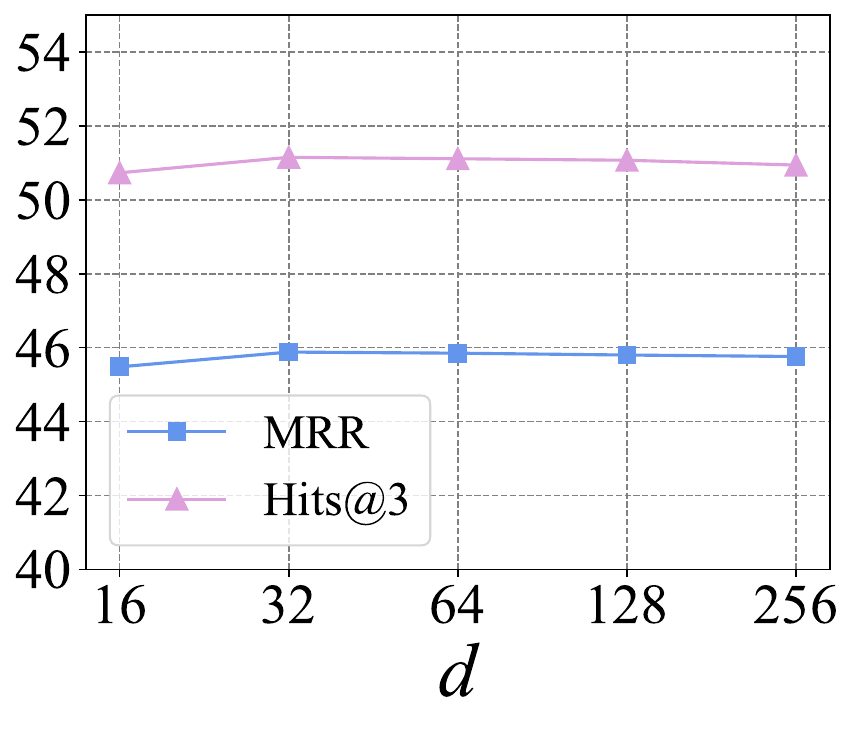}
		\end{minipage}
	}%
	\subfigure[ICE05-15]{
		\begin{minipage}[t]{0.48\linewidth}
			\centering
			\includegraphics[width=1\linewidth]{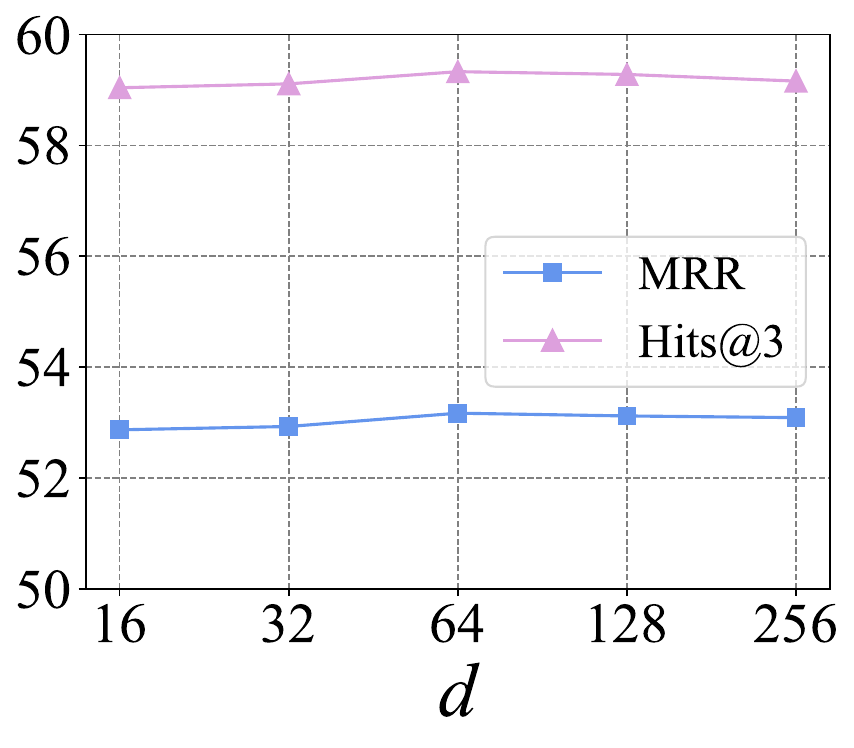}
		\end{minipage}
	}
	\caption{Study on different embedding size of relation $d$ on ICE14 and ICE05-15 datasets.}
	\label{fig:rel_size}
\end{figure}

\begin{figure}[htbp]
	\setlength{\abovecaptionskip}{0.5pt}
	\centering
	\subfigure[ICE14]{
		\begin{minipage}[t]{0.48\linewidth}
			\centering
			\includegraphics[width=1\linewidth]{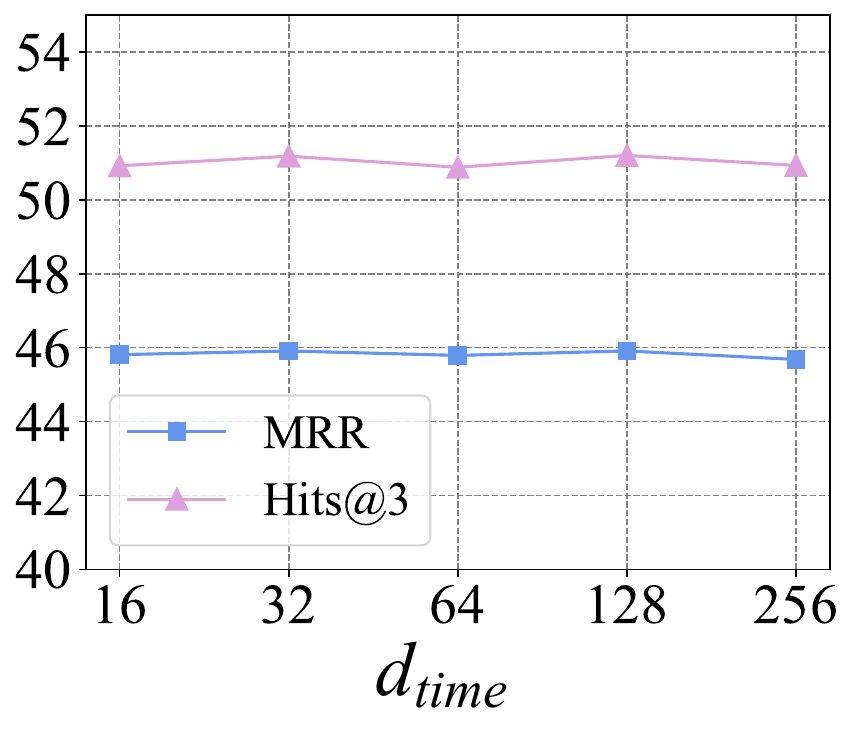}
		\end{minipage}
	}%
	\subfigure[ICE05-15]{
		\begin{minipage}[t]{0.48\linewidth}
			\centering
			\includegraphics[width=1\linewidth]{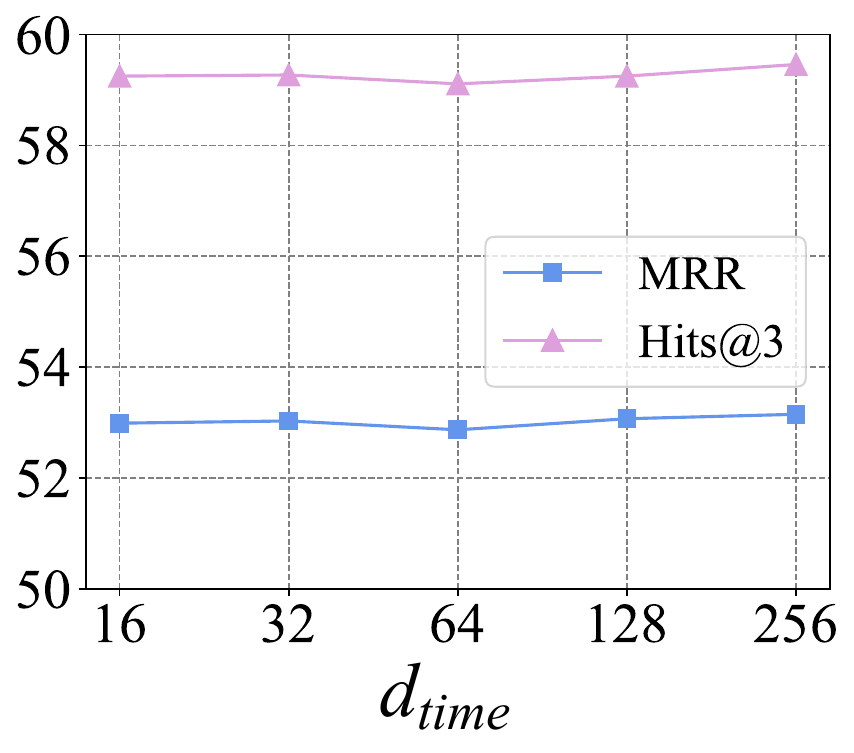}
		\end{minipage}
	}
	\caption{Study on different embedding size of time $d_{time}$ on ICE14 and ICE05-15 datasets.}
	\label{fig:time_size}
\end{figure}

Figure \ref{fig:rel_size} and \ref{fig:time_size} illustrate that altering the embedding size of relation $d$ and time $d_{time}$ exhibits minimal impact on the performance of CognTKE on ICE14 and ICE05-15 datasets, indicating the low sensitivity of the embedding size of relation and time. However, we can still intuitively see that the best performance is achieved when $d$ is set as 64 in Figure \ref{fig:rel_size} and $d_{time}$ is set as 32 in Figure \ref{fig:time_size}.

\subsection{Study of Prediction Time}
We compare the prediction time of CognTKE with that of the recent important extrapolation methods on all the datasets, including REGCN\footnote{\url{https://github.com/Lee-zix/RE-GCN}.}, CEN\footnote{\url{https://github.com/Lee-zix/CEN}.}, TLogic\footnote{\url{https://github.com/liu-yushan/TLogic}.}, TiRGN\footnote{\url{https://github.com/Liyyy2122/TiRGN}.}, CENET\footnote{\url{https://github.com/xyjigsaw/CENET}.} and RETIA\footnote{\url{https://github.com/CGCL-codes/RETIA}.}. To ensure a fair comparison, we used their open-source codes with optimal parameter settings for evaluation in the same environment. Table \ref{table:run_time} shows the results, with underlined values indicating better time efficiency than CognTKE.
CognTKE exhibits longer run times compared to REGCN, CEN, and TiGRN on all datasets due to the high time spent on TCR-Digraph construction in CognTKE. Tlogic is less time-efficient due to generate a large number of query-related paths. The time consumption of RETIA depends on the complex hyperrelational modeling employed. CENET's time efficiency is impacted by the significant time required for extracting historical repetitive entities and relations from the global historical space. 
In summary, CognTKE ensures only a limited increase in time efficiency, while providing excellent extrapolation results.
\begin{table}[h]
	\setlength\tabcolsep{2pt}
	\renewcommand{\arraystretch}{1}
	\caption{prediction time comparison of the recent state-of-the-art methods on all datasets (min: minutes; sec: seconds)}.
	\centering
	\scalebox{0.95}{
		\begin{tabular}{c|cccc}
			\toprule
			Dataset &ICE14  &ICE18 &ICE05-15 & WIKI     \\
			\midrule
			REGCN & \underline{4 sec} & \underline{23 sec} & \underline{1.33 min} & \underline{31 sec} \\
			CEN   & \underline{7 sec} & \underline{36 sec} & \underline{1.51 min} & \underline{32 sec} \\
			TLogic & 38.01 min & $>$24 hours & $>$24 hours &  - \\
			TiRGN & \underline{37 sec} & \underline{2.75 min} & \underline{10.46 min} & \underline{1.11 min}  \\
			CENET & 1.75 min & 19.48 min & 41.1 min & 18.75 min \\
			RETIA & 16.86 min & 49.7 min & 7.78 hours & \underline{41 sec} \\
			\bottomrule
			CognTKE & \textbf{47 sec} & \textbf{16.17 min} & \textbf{18.07 min} & \textbf{10.31 min}    \\
			\bottomrule 
		\end{tabular}
	}
	\setlength{\abovecaptionskip}{3pt}
	\label{table:run_time}
\end{table}

\subsection{C Proofs for Theorem 1}
\subsubsection{Proof.} According to Definition 4, temporal relation $p$ is a combination of relation $r$ and time $t$, the original quadruples are actually transformed into new triples. Any temporal relation paths $e^{t_q}_{q} \rightarrow p_{k}^1 \rightarrow p_{k}^2 \rightarrow \cdots \rightarrow p_{k}^L \rightarrow e^{t}_{o}$ in $\mathcal{C}$ are included in the TCR-Digraph  $\hat{\mathcal{G}}_{e^{t_q}_{q}, e^{t}_{o} \mid L}$.

The TCR-Digraph $\hat{\mathcal{G}}_{\mathcal{C}}$ is built by $\mathcal{C}$.  The attention weight of each layer is represented as:
\begin{equation}
	\begin{split}
		\mathbf{c}_{e_s,rt\mid r^{t_q}_{q}}^l&=\sigma_2 (\mathbf{W}_6^l\sigma_1(\mathbf{W}_3^l \mathbf{h}_{s}^l+  \mathbf{W}_4^l \mathbf{h}_{rt}^{l}+ \mathbf{W}_5^l \mathbf{h}_{r^{t_q}_{q}}^{l})) \\ &= \sigma_2 (\mathbf{W}_6^l\sigma_1(\mathbf{W}_3^l \mathbf{h}_{s}^l+  \mathbf{W}_4^l \mathbf{h}_{p}^{l}+ \mathbf{W}_5^l \mathbf{h}_{r^{t_q}_{q}}^{l}))\\
		&= f_{att}^{l}(e_s,p^l,e_q,r^{t_q}_{q}), \nonumber
	\end{split}
\end{equation}
where $p = rt$. Then, it needs to prove that  $\hat{\mathcal{G}}_{\mathcal{C}}$ can be recursively generated from the $L$-th layer to the first layer.

\begin{itemize}
	\item In the $L$-th layer, defined $\mathcal{D}_{+}^L$ as the set of the new triples $\left(e_{s}, p_k^L, e_{o}\right)$, where $p_k^L=(r_it_j)^L$, the attention weights of the new triples are $c_{e_s,rt\mid r^{t_q}_{q}}^L=f_{att}^{L}(e_s,p_k^L,e_q,r_q^{t_{q}})$, in the $L$-th layer of $\mathcal{G}_{\mathcal{C}}$. According to the universal approximation theorem, there exists a set of parameters $\mathbf{W}_3^L, \mathbf{W}_4^L, \mathbf{W}_5^L, \mathbf{W}_6^L, \mathbf{h}_{p}^{L}$,  that can learn a decision boundary $\theta$ that $\left(e_{s}, p_k^L, e_{o}\right) \in \mathcal{D}_{+}^L$ if $c_{e_s,rt\mid r^{t_q}_{q}}^L>\theta$ and otherwise $\left(e_{s}, p_k^L, e_{o}\right) \notin \mathcal{D}_{+}^L$. Then the $L$-th layer of $\hat{\mathcal{G}}_{\mathcal{C}}$ can be generated.
	
	\item Similarly, defined $\mathcal{D}_{+}^{L-1}$ as the set of the new triples $\left(e_{s}, p_k^{L-1}, e_{o}\right)$ that establishes connections with the remaining entities in the $L-1$-th layer. Then, there also exists a set of parameters $\mathbf{W}_3^{L-1}, \mathbf{W}_4^{L-1}, \mathbf{W}_5^{L-1}, \mathbf{W}_6^{L-1}, \mathbf{h}_{p}^{L-1}$ that can learn a decision boundary $\theta$ that $\left(e_{s}, p_k^{L-1}, e_{o}\right) \in$ $\mathcal{D}_{+}^{L-1}$ if $c_{e_s,rt\mid r^{t_q}_{q}}^{L-1}>\theta$ and otherwise not in. In addition, $\mathbf{W}_3^{L-1}, \mathbf{W}_4^{L-1}, \mathbf{W}_5^{L-1}, \mathbf{W}_6^{L-1}, \mathbf{h}_{p}^{L-1}$ and $\mathbf{W}_3^L, \mathbf{W}_4^L, \mathbf{W}_5^L, \mathbf{W}_6^L, \mathbf{h}_{p}^{L}$ are independent with each other. Thus, the $L-1$-th layer of $\hat{\mathcal{G}}_{\mathcal{C}}$ can be generated.
	
	\item At the last, $\hat{\mathcal{G}}_{\mathcal{C}}$ can be generated as the subgraph in $\hat{\mathcal{G}}_{e^{t_q}_{q}, e^{t}_{o} \mid L}$ with attention weights $c_{e_s,rt\mid r^{t_q}_{q}}^l$ by the recursive processing.
\end{itemize}

\end{document}